\pdfoutput=1

\documentclass[11pt]{article}

\usepackage[preprint]{acl}

\usepackage{times}
\usepackage{latexsym}
\usepackage{multirow}
\usepackage{booktabs}

\usepackage[T1]{fontenc}

\usepackage[utf8]{inputenc}

\usepackage{microtype}

\usepackage{inconsolata}

\usepackage{graphicx}
\usepackage{amsmath}
\usepackage{enumitem}
\usepackage{moresize}
\usepackage{geometry}
\usepackage{tikz} 
\usepackage{forest}
\usepackage{xcolor} 

\definecolor{harvestgold}{RGB}{227, 170, 68} 
\definecolor{carminepink}{RGB}{235, 76, 66} 
\definecolor{lightgreen}{RGB}{144, 238, 144} 
\definecolor{cyan}{RGB}{0, 255, 255} 
\definecolor{lightpurple}{RGB}{177, 156, 217} 

\newcommand\blfootnote[1]{%
\begingroup 
\renewcommand\thefootnote{}\footnote{#1}%
\addtocounter{footnote}{-1}%
\endgroup 
}

\newcommand{\ie}{\emph{i.e., }}
\newcommand{\eg}{\emph{e.g., }}

\newcommand{\cf}{\emph{cf. }}

\newcommand{\idea}{
  \begingroup\normalfont
  \includegraphics[height=1.3\fontcharht\font`\B]{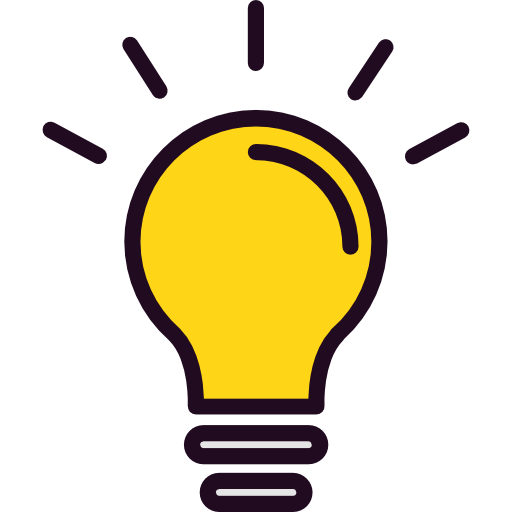}%
  \endgroup
}

\title{Knowledge Boundary of Large Language Models: A Survey}

\author{Moxin Li$^{1*}$ ~ Yong Zhao$^{12*\dagger}$ ~  \textbf{Wenxuan Zhang}$^3$ ~ \textbf{Shuaiyi Li}$^4$ ~ \textbf{Wenya Xie}$^5$ \\ \textbf{See-Kiong Ng}$^1$ ~ \textbf{Tat-Seng Chua}$^1$ ~ \textbf{Yang Deng}$^{2\ddagger}$ \\
        $^1$National University of Singapore \quad $^2$Singapore Management University \\ 
        $^3$Singapore University of Technology and Design\\ 
        $^4$The Chinese University of Hong Kong \quad $^5$University of Minnesota - Twin Cities \\ 
        \texttt{\{limoxin,yzhao\}@u.nus.edu}  \quad \texttt{ydeng@smu.edu.sg} }

\begin{document}
\maketitle
\begin{abstract}

Although large language models (LLMs) store vast amount of knowledge in their parameters, they still have limitations in the memorization and utilization of certain knowledge, leading to undesired behaviors such as generating untruthful and inaccurate responses. 
This highlights the critical need to understand the knowledge boundary of LLMs, a concept that remains inadequately defined in existing research.
In this survey, we propose a comprehensive definition of the LLM knowledge boundary and introduce a formalized taxonomy categorizing knowledge into four distinct types. 
Using this foundation, we systematically review the field through three key lenses: the motivation for studying LLM knowledge boundaries, methods for identifying these boundaries, and strategies for mitigating the challenges they present. 
Finally, we discuss open challenges and potential research directions in this area. 
We aim for this survey to offer the community a comprehensive overview, facilitate access to key issues, and inspire further advancements in LLM knowledge research.
\blfootnote{$^*$ Equal contribution.}
\blfootnote{$^\dagger$ Work was done during an internship at SMU.}
\blfootnote{$^\ddagger$ Corresponding author.}

\end{abstract}

\section{Introduction}

Large language models (LLMs) store extensive knowledge within their parameters, enabling impressive performance across a wide range of tasks. 
However, LLMs have been criticized for significant issues related to the memorization and utilization of knowledge, such as generating responses that contain untruthful information \cite{csur-hallucination}, being misled by untruthful context \cite{wang2023can}, or lacking precision to unclear queries \cite{Zhang2024CLAMBERAB}. 
In light of this, recent studies have introduced the concept of LLM knowledge boundary \cite{DBLP:conf/acl/YinZR024}, 
defining knowledge types based on the LLM's performance in knowledge question answering (QA). 
Understanding the knowledge boundary is crucial for ensuring the trustworthy deployment of LLMs.

We identify the major limitations in existing definitions of the LLM knowledge boundary. 
Firstly, the Know-Unknow Quadrant \cite{DBLP:conf/acl/YinSGWQH23, DBLP:conf/acl/AmayuelasWPCW24, Li2025RefineKO} categorizes knowledge based on the LLM's possession and the LLM's awareness of such knowledge, but this definition is conceptual and lacks formalization. 
Besides, \citet{DBLP:conf/acl/YinZR024} introduce a formalized definition separating the influence of the prompt from the LLM's mastery of the knowledge, yet they merely focus on the knowledge boundary of a specific LLM which lacks comprehensiveness. 
Additionally, some recent surveys \cite{li2024survey,wen2024know} also discuss certain topics related to the LLM knowledge boundary. However, \citet{li2024survey} lack a clear and formalized definition, and \citet{wen2024know} merely focus on the abstention strategy for handling knowledge limitation. 
These limitations hinder a thorough and nuanced understanding of the LLM knowledge boundary.

\begin{figure*}
    \centering
    \includegraphics[width=0.98\textwidth]{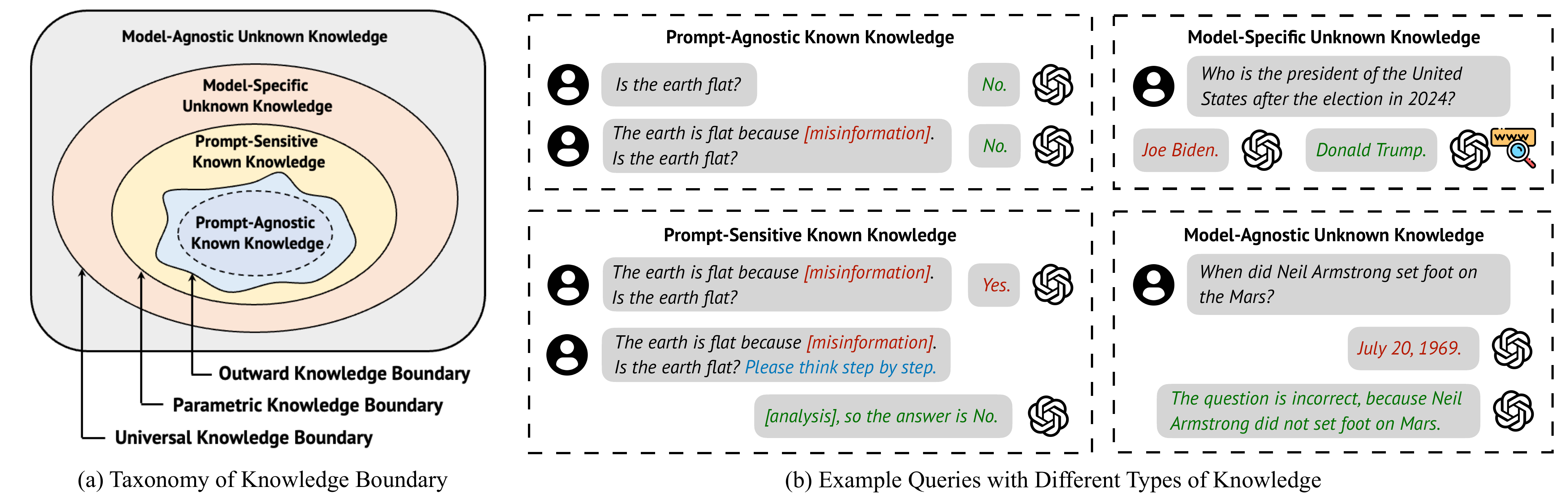}
    \caption{Illustration of the knowledge boundaries and knowledge taxonomy of LLM. The dashed circle in (a) represents the ``truly'' prompt-agnostic known knowledge $k$, which can be verified by any expression in $Q_k$. In practice, however, the prompt-agnostic nature of $k$ can only be approximated using a limited subset $\hat{Q}_k\subseteq Q_k$. As a result, the outward knowledge boundary is depicted with an irregularly shaped line to reflect this approximation. }
    \label{fig:knowledge_boundary_def}
\end{figure*}

To address these limitations, we propose a comprehensive and formalized definition of the knowledge boundary of LLMs. 
Our definition classifies knowledge from three dimensions: 1) whether the knowledge is known to human and expressible in textual QA form (\textbf{\textit{Universal Knowledge Boundary}}), 
2) whether it is abstractly embedded within the LLM’s parameters (\textbf{\textit{Parametric Knowledge Boundary}}), 
and 3) whether it is empirically validated on the LLM (\textbf{\textit{Outward Knowledge Boundary}}). 
Based on these knowledge boundaries, we establish a formal four-type knowledge taxonomy to classify and define each knowledge type (\S~\ref{sec:preliminary}).

Building on our proposed taxonomy, we systematically review related research. 
Our survey is organized around three key research questions. 
First, we address \textbf{\textit{RQ1: Why study knowledge boundaries?}}, by detailing the LLMs' undesirable behaviors that stem from their unawareness of knowledge boundaries (\S~\ref{sec:undesired_behaviours}). 
Next, we explore \textbf{\textit{RQ2: How can knowledge boundaries be identified?}}, highlighting uncertainty, calibration and probing techniques to distinguish different knowledge types (\S~\ref{sec:identification}). 
Furthermore, we investigate \textbf{\textit{RQ3: How can issues caused by knowledge boundaries be mitigated?}}, summarizing strategies to enhance the task performance and foster desired behaviors for each knowledge type (\S~\ref{sec:mitigation}).

Finally, we discuss the open challenges and prospective directions for advancing the understanding of the LLM knowledge boundary (\S~\ref{sec:challenge}). 
First, we advocate for more comprehensive benchmarks to assess knowledge boundaries across various types of knowledge limitations. Second, we emphasize the potential utilization of LLM knowledge boundaries in future  developments of LLMs.  
Lastly, we discuss the role of knowledge boundary in different knowledge mechanisms.

The overview of this survey\footnote{\url{https://github.com/li-moxin/knowledge_boundary_survey}.} and related datasets are presented in Appendix \ref{app:overview} and \ref{app:dataset}, respectively.

\section{Definition of Knowledge Boundary} \label{sec:preliminary}
To mitigate the shortcomings of existing definitions, we provide a more complete and formalized definition of the knowledge boundary for LLMs. 
Formally, we denote $\mathcal{K}$ as the whole set of abstracted knowledge that is known to human, and $k$ as a piece of knowledge that can be expressed by a set of input-output pairs $Q_k = \{(q_k^i,a_k^i)\}_i$.
Let $\theta$ represent the parameters of a specific LLM. 
As shown in Figure~\ref{fig:knowledge_boundary_def}, we define three types of knowledge boundaries for LLMs where one subsumes another:
\begin{itemize}[leftmargin=*,nosep]
    \item \textbf{\textit{Outward Knowledge Boundary}} defines the observable knowledge boundary for a specific LLM. 
    The knowledge verification is usually conducted on a limited available subset of expressions $\hat{Q}_k \subseteq Q_k$.
    Knowledge within this boundary refers to the knowledge that the LLM can generate correct outputs for the input for all instances in $\hat{Q}_k$. 
    \item \textbf{\textit{Parametric Knowledge Boundary}} defines the abstract knowledge boundary for a specific LLM. Knowledge within this boundary is possessed in the LLM parameters, which could be verified by at least one expression in $Q_k$.
    \item \textit{\textbf{Universal Knowledge Boundary}} defines the whole set of knowledge known to human, which is verifiable by certain input-output pairs in $Q_k$. 
\end{itemize}
\vspace{5pt}
Divided by the knowledge boundaries, four types of knowledge are defined as below. Figure \ref{fig:knowledge_boundary_def} presents example queries with each type of knowledge. 
\begin{itemize}[leftmargin=*,nosep]
    \item \textbf{Prompt-Agnostic Known Knowledge} (\texttt{PAK}) can be verified by all expressions in $\hat{Q}_k$ for the LLM $\theta$ regardless of the prompt, \ie the predicted output probability is larger than a threshold $\epsilon$. 
    \begin{equation}\small
    K_\texttt{PAK} = \{k\in\mathcal{K}|\forall (q_k^i, a_k^i)\in \hat{Q}_k, P_{\theta}(a_k^i|q_k^i)>\epsilon\}
    \end{equation}

    \item \textbf{Prompt-Sensitive Known Knowledge} (\texttt{PSK}) resides within the LLM's parameters but is sensitive to the form of the prompt. While certain expressions in $\hat{Q}_k$ may fail to verify this type of knowledge, appropriate expressions in $Q_k$ can be found for successful verification.  
    \begin{equation}\small\begin{split}
    K_\texttt{PSK} = \{k\in\mathcal{K}| (\exists (q_k^i, a_k^i)\in Q_k, P_{\theta}(a_k^i|q_k^i)>\epsilon) \\\wedge (\exists (q_k^i, a_k^i)\in \hat{Q}_k, P_{\theta}(a_k^i|q_k^i)<\epsilon)\}
    \end{split}\end{equation}

    \item \textbf{Model-Specific Unknown Knowledge} (\texttt{MSU}) is not possessed in the specific LLM parameters $\theta$, thus cannot be verified by any instance in $Q_k$ for the LLM, but the knowledge itself is known to human, \ie $Q_k$ is non-empty. 
    \begin{equation}\small
    K_\texttt{MSU} = \{k\in\mathcal{K}| \forall (q_k^i, a_k^i)\in Q_k, P_{\theta}(a_k^i|q_k^i)<\epsilon\}
    \end{equation}

    \item \textbf{Model-Agnostic Unknown Knowledge} (\texttt{MAU}) is unknown to human (\textit{i.e.}, $Q_k$ is empty), thus unverifiable regardless of the model. 
    \begin{equation}\small
    K_\texttt{MAU} = \{k\in\mathcal{K}| Q_k = \text{\O} \}
    \end{equation}
\end{itemize}

\noindent \colorbox{violet!8}{
\begin{minipage}{0.46\textwidth}
\ssmall
{\color{violet}{\textbf{Summary \& Ideas - Definition of Knowledge Boundary}}}
\begin{itemize}[itemsep=-1pt, topsep=2pt, leftmargin=6pt, labelsep=3pt]
\item We provide a formalized definition for LLM knowledge boundaries, and define a four-type knowledge taxonomy accordingly.  
\item Our knowledge taxonomy can also be adapted to the Know-Unknow Quadrant \cite{DBLP:conf/acl/YinSGWQH23,DBLP:conf/acl/AmayuelasWPCW24}, where \texttt{PAK} and \texttt{PSK} can be viewed as a form of the known-knowns and the unknown-knowns respectively, while \texttt{MSK} and \texttt{MAK} jointly formulate the known-unknowns. 
\setlength{\labelsep}{1pt}
\item[\idea] We do not explicitly define the unknown-unknown, since it is largely underexplored in the study of LLM knowledge. Future research can further explore the unknown-unknowns for LLMs and humans. 
\end{itemize}
\end{minipage}
} \\ [-2mm]

\section{Undesired Behaviours} \label{sec:undesired_behaviours}
We first address \textbf{\textit{RQ1: Why study knowledge boundaries?}}
Due to the unawareness of knowledge boundaries, LLMs exhibit various undesired behaviors that compromise the reliability and utility of their outputs, 
posing challenges for the successful applications of LLMs. In this section, we delineate three types of undesired behaviours according to the unawareness of each knowledge boundary.

\subsection{Untruthful Responses Misled by Context}
Even though LLMs possess the required knowledge, they often produce untruthful responses when misled by context, \ie the prompt-sensitive known knowledge. 
This can occur in two forms: \textit{untruthful context}, where the context includes false or misleading information, and \textit{irrelevant context}, where extraneous details divert the model from generating precise responses.

\vspace{-3pt}
\paragraph{Untruthful Context}
Incorporating false information into the context significantly biases LLMs, severely impacting their performance \cite{chen2024editingllmsinjectharm, Pan2023OnTR}. Using in-context learning (ICL) allows for editing factual knowledge in LLMs, which may lead to varied factual outputs \cite{ike}. When faced with untruthful views, LLMs often fail to stay true \cite{aaai25-integrity}, being swayed by persuasive tactics despite initially correct responses \cite{wang2023can, Xu2023TheEI}.

\vspace{-3pt}
\paragraph{Irrelevant Context}
Irrelevant context can dramatically affect LLMs, leading to off-topic or inaccurate responses. Irrelevant details in problem descriptions or retrieval systems drastically undermine model performance \cite{shi2023large}. When such information is semantically related to the context, it exacerbates this effect, causing LLMs to overlook crucial information and reduce response accuracy \cite{Wu2024HowED}.

\subsection{Factuality Hallucinations}\label{sec:hallucination}

\textit{Factuality hallucinations} \cite{Huang2023ASO}, \textit{i.e.}, the model output diverges from real-world facts, typically stem from the following causes, relating to the model-specific unknown knowledge. 

\vspace{-3pt}
\paragraph{Deficiency of Domain-specific Knowledge}
LLMs, primarily trained on broad, publicly accessible datasets, often lack detailed knowledge in specialized domains, leading to inaccuracies in domain-specific queries. For example, ChatGPT often issues incorrect or imprecise biomedical advice \cite{pal2024domain}, and misrepresents legal facts or arguments \cite{legal-hallucination}. Similar issues arise in medical \cite{medical-hallucination} and financial contexts \cite{finance-hallucination}, where LLMs exhibit hallucinations due to insufficient domain-specific knowledge.

\vspace{-3pt}
\paragraph{Outdated Knowledge}
A significant limitation of LLMs is their reliance on outdated information, as their training data is bounded by temporal limitations. Without mechanisms to update their internal knowledge, LLMs struggle to adapt to new developments, often resorting to fabricating facts or using outdated responses \cite{naacl22-outdated,kasai2024realtime}. For instance, LLaMA2 \cite{DBLP:journals/corr/abs-2307-09288}, despite its recent training cutoffs (e.g., 2022), tends to use data from earlier years (e.g., 2019) \cite{Zhao2024SetTC}. Recent studies like \citet{Cheng2024DatedDT} highlight these temporal knowledge cutoffs, revealing the scope of outdated information in LLMs.

\vspace{-3pt}
\paragraph{Overconfidence on Unknown Knowledge}
LLMs often show overconfidence when addressing topics beyond their knowledge, delivering assertive but incorrect responses. This tendency is partly due to the limited generalization of their reward systems which overfit familiar data and neglect less-known subjects, thus leading to amplifying overconfident outputs \cite{yan2024reward}. LLMs also lack mechanisms to indicate uncertainty or acknowledge knowledge limits, which exacerbates the issue of overconfidence.  Studies have shown that LLMs perform poorly on unfamiliar topics while maintaining high confidence \cite{agarwal2023nlpmodelsidentifydistinguish, deng2024gotcha}.

\subsection{Truthful but Undesired Responses}
LLMs may produce improper responses when handling model-agnostic unknown knowledge, leading to answers misaligned with user expectations.

\vspace{-3pt}
\paragraph{Random Responses to Ambiguous Knowledge}

Ambiguous knowledge challenges LLMs' understanding, often leading them to guess responses due to their inability to recognize ambiguities \cite{emnlp23-ambiguity, Zhang2024CLAMBERAB}. 
They typically provide arbitrary answers to unclear queries \cite{Deng2023PromptingAE}, 
or generate a mix of low-probability correct answers and incorrect answers to semi-open-ended queries \cite{DBLP:journals/corr/abs-2405-14383}.

\vspace{-3pt}
\paragraph{Biased Responses to Controversial Knowledge}

Controversial knowledge involves subjective questions with varied answers depending on individual perspectives \cite{coling24-controversial,DBLP:conf/acl/AmayuelasWPCW24}. These reveal biases in LLMs trained on skewed datasets, leading to partiality in responses. Such bias may cause unfair emphasis on certain viewpoints or stereotypical portrayals of demographics, exacerbating disparities \cite{Singh2024BornWA,acl24-culturalbias}.

\vspace{2mm}
\noindent \colorbox{violet!8}{
\begin{minipage}{0.46\textwidth}
\ssmall
{\color{violet}{\textbf{Summary \& Ideas - Undesired Behaviors}}}
\begin{itemize}[itemsep=-1pt, topsep=2pt, leftmargin=6pt, labelsep=3pt]
\item Due to the unawareness of knowledge boundaries, LLMs often exhibit factuality hallucinations caused by outdated or insufficient domain knowledge and overconfidence on unknown knowledge, are susceptible to being misled by untruthful or irrelevant context, and produce random or biased responses that don't align with user expectations. 
\setlength{\labelsep}{1pt}
\item[\idea] 
Despite their strong relevance to the knowledge boundary of LLMs, existing studies fail to analyze or address these undesired behaviours through the lens of knowledge boundary, which can provide insights into their underlying causes and help develop strategies to mitigate their impact.
\end{itemize}
\end{minipage}
} \\ [-2mm]

\section{Identification of Knowledge Boundary} \label{sec:identification}

We then delve into \textbf{\textit{RQ2: How to identify knowledge boundaries?}}
We categorize the existing solutions into three types:
\emph{uncertainty estimation}, \emph{confidence calibration}, and \emph{internal state probing}.

\subsection{Uncertainty Estimation}
Uncertainty estimation (UE) aims to quantify the uncertainty of a model regarding its predictions for a given input. 
High uncertainty indicates that the model is unlikely to produce correct predictions to the input, thus the input-related knowledge lies outside of certain knowledge boundaries of the model. 
UE has been widely studied on NLP models \cite{hu2023uncertainty}. 
In the era of LLMs, we highlight the following four groups of studies. 

\vspace{-3pt}
\paragraph{Uncertainty Decomposition}
The uncertainty of LLM can be decomposed into \textit{epistemic uncertainty} and \textit{aleatoric uncertainty} \cite{hou2023decomposing}. 
\textit{Epistemic uncertainty} refers to the model-specific uncertainty, quantifying the lack of model knowledge, which is related to our definition of \textbf{\textit{Parametric Knowledge Boundary}}. 
\textit{Aleatoric uncertainty} refers to the data-level uncertainty, such as ambiguous prompts having multiple valid answers, referring to the gap between \textbf{\textit{Outward Knowledge Boundary}} and \textbf{\textit{Parametric Knowledge Boundary}}. 
Quantifying these types of uncertainty can help to identify different approaches for mitigating the knowledge limitations (Section~\ref{sec:mitigation}). 
Solutions to quantify the two types of uncertainty can be roughly classified into data-side and model-side approaches, where one type of uncertainty can be obtained by subtracting the other type from the total uncertainty. 
The data-side quantification include input-side clarification and perturbation \cite{hou2023decomposing, ling2024uncertainty, gao2024spuq}, and output-side variation estimation \cite{yadkori2024believe, aichberger2024how}. 
The model-side quantification include model parameter and configuration perturbation \cite{ling2024uncertainty} and model internal states perturbation \cite{ahdritz2024distinguishing}.

However, many other current approaches of UE do not distinguish the two types of uncertainty and focus on the general identification of the \textbf{\textit{Outward Knowledge Boundary}}, detailed as below.

\paragraph{Conformal Prediction}
Conformal Prediction \cite{DBLP:journals/sigact/Law06} quantifies the uncertainty of model outputs by identifying a set of outputs with a guaranteed probability that the correct output is included within the set. 
This approach offers advantages such as being logit-free and suitable for black-box LLMs \cite{DBLP:conf/emnlp/SuLWC24}. Several studies have explored and attempted to address the issue of overconfidence in LLM conformal prediction \cite{DBLP:conf/acl/RavfogelGG23, DBLP:conf/nips/0001YPWWY0T24}. Furthermore, conformal prediction has been applied in techniques such as prompt selection \cite{DBLP:conf/iclr/ZolloMDSPZ24}, decoding stopping rules for guaranteed generation \cite{DBLP:conf/iclr/QuachFSYSJB24}, and ensuring reliability in retrieval-augmented generation \cite{DBLP:conf/naacl/Li00B24}.

\vspace{-3pt}
\paragraph{Token Probability-based Uncertainty Estimation}
Stemming from the traditional UE, the straightforward token probability-based UE computes the average token probability or the entropy of the LLM predictions as the uncertainty \cite{manakul2023selfcheckgpt, huang2023look}. 
Detailed designs involve considering different granularities of the predictions beyond token-level, such as sentence-level \cite{duan2024shifting} and atomic fact-level \cite{fadeeva2024fact}, weighted by the relevance of different components \cite{duan2024shifting}.

\vspace{-3pt}
\paragraph{Semantic-based Uncertainty Estimation}
The token probability-based UE are unsuitable for proprietary LLMs, and might be insufficient in quantifying the semantic uncertainty of LLM predictions. 
Therefore, the semantic-based UE is proposed, roughly categorized into \textit{consistency-based methods} and \textit{verbalized methods}. 
The \textit{\textbf{consistency-based methods}} view the inconsistency among multiple sampled predictions of the input as the uncertainty. 
The approaches to measure the semantic consistency of the sampled outputs include 
the semantic distance calculated by smaller models \cite{kuhn2023semantic, lin2024generating, zhao2024knowing, nikitin2024kernel, manakul2023selfcheckgpt}, and the consistency in the LLM evaluation \cite{chen2024quantifying, manakul2023selfcheckgpt}. 
The \textit{\textbf{verbalized methods}} aim to enable LLMs to express their uncertainty directly as output tokens. 
\citet{zhou2024relying} reveal that LLMs are reluctant to verbally express their uncertainty, possibly related to the lack of uncertainty expression in the training data.
\citet{lin2022teaching} and \citet{chaudhry2024finetuning} adopt ICL and fine-tuning approaches to teach LLMs to generate uncertainty expressions.

\subsection{Confidence Calibration}
Calibration refers to the alignment between the estimated LLM confidence and the actual prediction correctness. This type of approach evaluates the confidence level of the LLM in a certain prediction. 
Low confidence suggests potential inaccurate prediction, indicating that the LLM may lack certain knowledge. We categorize existing methods into \textit{prompt-based} and \textit{fine-tuning} approaches. 

\paragraph{Prompt-based Calibration}

One group of approaches aims to prompt LLMs to \textbf{elicit confidence}, according to the prediction probability as a measure of the LLM confidence via sampling \cite{si2023prompting, wang2023selfconsistency}, or by the probability of the prediction being evaluated as correct by  LLMs \cite{kadavath2022language}. 
Techniques to improve calibration include prompt ensemble \cite{jiang2023calibrating}, hybrid approach \cite{chen2024quantifying}, fidelity evaluation \cite{zhang2024calibrating}, and model ensemble \cite{shrivastava2024llamas, feng-etal-2024-dont}.

Another group of approaches aims to prompt LLMs to directly \textbf{express confidence} as tokens in the prediction. 
Prompting RLHF-LLMs to express confidence can achieve better calibration than using token probability \cite{tian2023just}, and prompting LLMs to generating explanations can further be leveraged to enhance calibration \cite{zhao-etal-2024-fact, li2024think}. Combination with the former prompting approach can further improve performance \cite{xiong2024can}. 

\paragraph{Fine-tuning for Calibration}
The fine-tuning methods involve self-updating the LLM parameters and tuning additional models for calibration. 
The self-update involves instruction tuning for confidence expression \cite{tao-etal-2024-trust}, and learning to adjust the output token probabilities \cite{liu2024litcab, xie-etal-2024-calibrating}. 
Additional models can be trained for adjusting the LLM output probability towards calibration \cite{shen2024thermometer}, or directly evaluating the correctness and estimating the confidence level of the LLM outputs \cite{mielke2022reducing, stengel2024lacie}.

\subsection{Internal State Probing} \label{sec:internal_state_probing}
The internal states of LLM contain information related to the knowledge boundary. 
Linear probing on the internal states can be used to assess the factual accuracy of the LLM predictions \cite{li2024inference, azaria-mitchell-2023-internal, burns2022discovering, kossen2024semantic}, thus detecting the knowledge boundaries. 
The internal states involve attention heads \cite{li2024inference}, hidden layer activations \cite{azaria-mitchell-2023-internal, ji-etal-2024-llm, burns2022discovering}, neurons and tokens \cite{ji-etal-2024-llm}. \citet{marks2023geometry} validate the rationality of the linear probes. Moreover, \citet{liu2024universal} and \citet{marks2023geometry} study the the generalization ability of the probing method.

\vspace{1mm}
\noindent \colorbox{violet!8}{
\begin{minipage}{0.46\textwidth}
\ssmall
{\color{violet}{\textbf{Summary \& Ideas - Identification of Knowledge Boundary}}}
\begin{itemize}[itemsep=-1pt, topsep=2pt, leftmargin=6pt, labelsep=3pt]
\item Most of the existing identification approaches target at the the outward knowledge boundary, while the uncertainty decomposition is also concerned about the parametric knowledge boundary. 
\item Uncertainty estimation (UE) and confidence calibration are similar concepts but different in that confidence calibration targets at certain predictions, while UE aims for the entire prediction distribution \cite{DBLP:journals/corr/abs-2410-15326, wen2024know}. 
\setlength{\labelsep}{1pt}
\item[\idea] Identification approaches should be designed for different knowledge boundaries, suiting different mitigation approaches. 
\end{itemize}
\end{minipage}
} \\ [-2mm]

\section{Mitigation} \label{sec:mitigation}

Following the identification of knowledge boundaries, we discuss \textit{\textbf{RQ3: How to mitigate the issues caused by the knowledge boundaries?}} This section is organized following our knowledge taxonomy.

\subsection{Prompt-sensitive Known Knowledge}
\begin{figure}
    \centering
    \includegraphics[width=\linewidth]{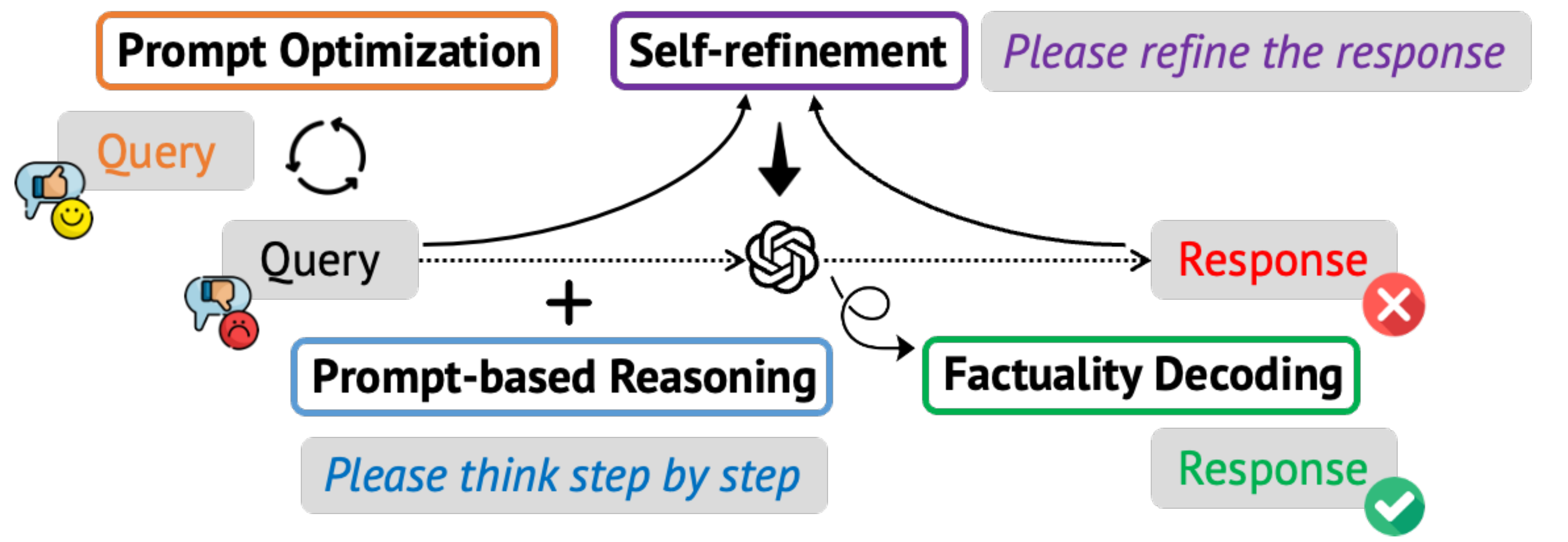}
    \caption{Summary of the mitigation techniques for prompt-sensitive known knowledge. }
    \label{fig:mitigation_5.1}
\end{figure}

The undesired outputs for this type of knowledge stem from inappropriate user prompts that fail to activate the embedded knowledge within the LLM. Accordingly, mitigation strategies typically focus on crafting suitable prompts to better leverage the LLM's knowledge, thereby improving task performance. 
We introduce four types of approaches as summarized in Figure~\ref{fig:mitigation_5.1}. 

\paragraph{Prompt Optimization}

Optimizing the prompt phrasing is essential for the LLM knowledge utilization and improved task performance. 
This approach can be categorized into two areas: \textit{instruction optimization} and \textit{demonstration optimization}. 

For instruction optimization, training-free methods include search-based techniques like Monte Carlo search \cite{zhou2023large, li2023robust, yang-etal-2024-dual}, tree search \cite{wang2024promptagent}, and searching on edit operations \cite{prasad2023grips}, where the LLM is often involved as the prompt optimizer \cite{yang2024large, pryzant2023automatic, do2023prompt}. 
The training-based methods typically rely on reinforcement learning to train additional modules for prompt optimization \cite{zhang2023tempera, deng2022rlprompt, diao2023blackbox}. 

For demonstration optimization, the diversity and similarity of the demonstrations are crucial factors for optimization \cite{xu2024context}. 
For example, the similar demonstrations are found by K-Nearest Neighbors \cite{liu2022makes} and BM25 \cite{luo2023dricl}, and the diverse demonstrations are identified by support example selection \cite{li2023finding} and diversity sampling \cite{mavromatis2023examples}. 
Effective demonstrations can also be identified by training ranking models according to better LLM task performance \cite{li2023unified, rubin2022learning, iter2023context, ye2023compositional}.

\paragraph{Prompt-based Reasoning}

Prompt-based reasoning strategies are often adopted to improve the LLM knowledge utilization \cite{wei2022chain, DBLP:conf/iclr/ZhouSHWS0SCBLC23, DBLP:conf/nips/YaoYZS00N23, DBLP:journals/corr/abs-2304-09797}.
For multi-step knowledge-based QA, the process generally involves individual steps such as question decomposition \cite{press2023measuring}, knowledge elicitation and inference \cite{wang2022iteratively, jung2022maieutic, liu2022generated}. 
External knowledge is often involved in this process to mitigate the knowledge gaps \cite{zhang2024tree, wu2024gendec, zhao2023verify, li2024chainofknowledge}.

\paragraph{Self-refinement}

The iterative self-refinement of the initial LLM prediction is also beneficial for knowledge utilization. 
The approaches can be broadly divided into \textit{single-model refinement} and \textit{multi-agent debate}.
For single-model refinement, LLMs are prompted to refine the predictions under a designed evaluation and regeneration process \cite{madaan2024self, miao2024selfcheck}, or generate self-verification questions to check for prediction consistency \cite{manakul2023selfcheckgpt, weng2023large}.  
While \citet{huang2024large} critique that LLMs struggle to achieve self-refinement without external feedback, \citet{li2024confidence} show that self-estimated confidence may improve self-refinement. 
In multi-agent debate, the LLM plays different roles to assess and refine its predictions from multiple angles \cite{du2024improving, fu2023improving}.

\paragraph{Factuality Decoding}
Different decoding strategies can also affect the LLM knowledge utilization, thus affecting the prediction factuality, which falls into two categories \cite{bi2024factuality}. 
The first category involves contrastive decoding against naive predictions with potential factual errors \cite{kim-etal-2024-adaptive}. 
The predictions for contrast come from smaller LLMs \cite{li2023contrastive}, lower layers of the LLM \cite{chuang2024dola, chen2024lower}, tokens with lower predicted probabilities \cite{kai2024sh2}, or predictions with induced hallucination \cite{yang2024improving, zhang2023alleviating}. 
The second category leverages the truthful directions identified from LLM internal states (\S~\ref{sec:internal_state_probing}). By editing these internal representations during decoding, it steers the model towards truthful directions, thereby enhancing the factuality of predictions \cite{li2024inference, chen2024context, qiu2024spectral, chen2024truth, zhang2024truthx}.

\vspace{5mm}
\noindent \colorbox{violet!8}{
\begin{minipage}{0.46\textwidth}
\ssmall
{\color{violet}{\textbf{Summary \& Ideas - Mitigation of Prompt-sensitive Known Knowledge}}}
\begin{itemize}[itemsep=-1pt, topsep=2pt, leftmargin=6pt, labelsep=3pt]
\item Improving the utilization of prompt-sensitivity known knowledge can be achieved from both the LLM input and output sides (\cf Figure~\ref{fig:mitigation_5.1}). 
\setlength{\labelsep}{1pt}
\item[\idea] A potential research gap lies in reducing the prompt sensitivity of LLMs. Future research can focus on the possibility and rationality of reducing the prompt sensitivity towards effective LLM knowledge utilization.
\end{itemize}
\end{minipage}
} 
\subsection{Model-specific Unknown Knowledge}

\begin{figure}
    \setlength{\abovecaptionskip}{5pt}   
    \setlength{\belowcaptionskip}{0pt}
    \centering
    \includegraphics[width=\linewidth]{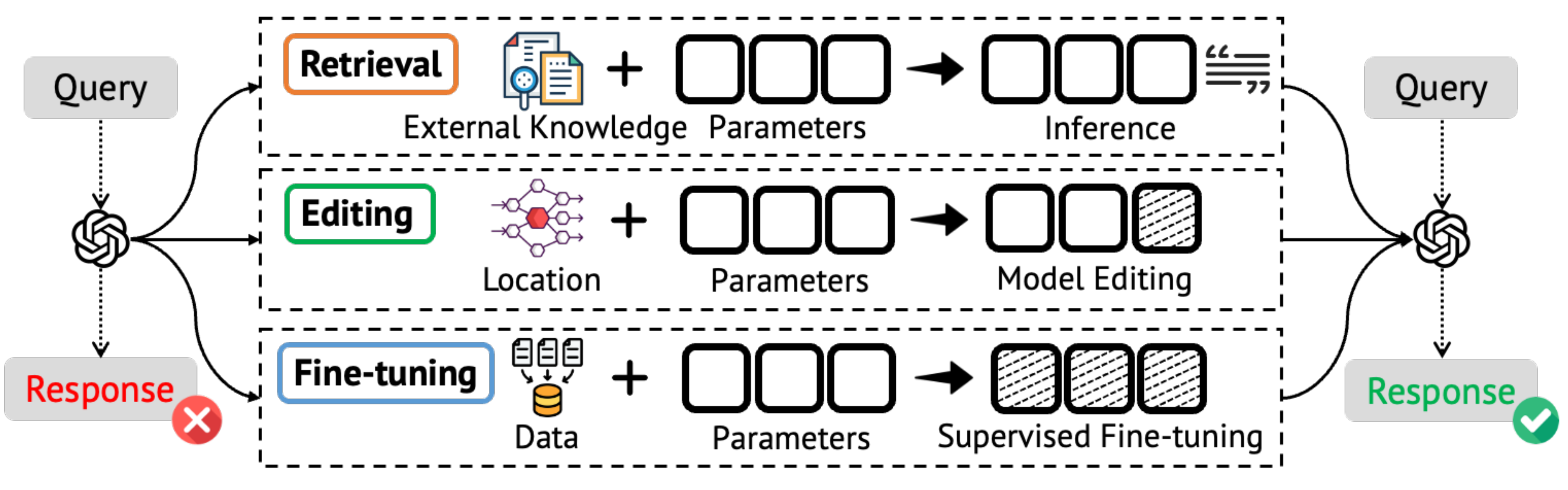}
    \caption{Summary of the mitigation techniques for model-specific unknown knowledge. }
    \label{fig:mitigation_5.2}
    \vspace{-3mm}
\end{figure}

The mitigation of model-specific unknown knowledge focuses on bridging gaps in domain-specific or up-to-date knowledge that fall outside the models' training data. Figure \ref{fig:mitigation_5.2} illustrates the mitigation strategies categorized into three key approaches.

\paragraph{External Knowledge Retrieval}
External knowledge retrieval is typically used for retrieval-augmented generation (RAG), which dynamically incorporates external knowledge during inference, expanding the effective knowledge boundary of LLMs~\cite{ren2023investigating}. Existing approaches can be divided into \textit{pre-generation} and \textit{on-demand} retrieval methods.
\textit{\textbf{Pre-generation}} methods ~\cite{gao-etal-2023-precise, shi-etal-2024-replug, yang-etal-2023-prca, wang2023learning} enhance the accuracy and relevance of responses by optimizing the retrieval process through methods such as refining user queries \cite{gao-etal-2023-precise,ma-etal-2023-query}, leveraging reader performance signals \cite{shi-etal-2024-replug}, and incorporating intermediary components that better align the retrieved knowledge with the knowledge needs of LLM \cite{yang-etal-2023-prca,ke-etal-2024-bridging,wang2023learning}.
\textit{\textbf{On-demand}} techniques adaptively retrieve external knowledge during generation, based on the LLM's confidence on its responses~\cite{jiang-etal-2023-active}, self-reflection results \cite{asai2024selfrag}, or iterative retrieval \cite{shao-etal-2023-enhancing}. The goal is to refine the interaction between retrieved and parametric knowledge while mitigating factual gaps.  

\paragraph{Parametric Knowledge Editing}

Researchers also develop knowledge editing methods for altering model behaviors to modify specific parameters within the LLM without extensive retraining. 
According to the memory mechanism, we categorize existing knowledge editing methods into three categories: \textit{explicit memory space}, \textit{implicit memory space}, and \textit{no memory space}. 
As for \textit{explicit memory space}, these approaches \cite{serac,ike,memprompt,bbox-edit,mello} use a memory pool to retrieve and apply edits via prompts. 
As for \textit{implicit memory space}, these approaches activate the LLM's parametric memory space based on specific input triggers, such as codebook \cite{grace}, neurons \cite{t-patcher,calinet}, LoRA blocks \cite{melo}, and FFN side memories \cite{wise}.
Another group of methods does not adopt extra memory components. Instead, they adopt various techniques to directly edit the original model parameters, such as meta learning \cite{malmen} and locate-then-edit \cite{rome,memit}.

\paragraph{Knowledge-enhanced Fine-tuning}

Knowledge-enhanced fine-tuning internalizes new knowledge into models by leveraging structured or synthetic representations. This involves encoding knowledge as factual records, synthetic corpora, and domain-specific taxonomies. Techniques such as fact-based encoding \cite{Mecklenburg2024InjectingNK}, synthetic data creation \cite{Joshi2024AdaptingML}, and hierarchical organization \cite{Liu2024StructureawareDK} ensure comprehensive domain coverage, while interleaved generation and context-aware structuring \cite{Zhang2024SyntheticKI} aim to enhance the data quality.

\vspace{1mm}
\noindent \colorbox{violet!8}{
\begin{minipage}{0.46\textwidth}
\ssmall
{\color{violet}{\textbf{Summary \& Ideas - Mitigation of Model-specific Unknown Knowledge}}}
\begin{itemize}[itemsep=-1pt, topsep=2pt, leftmargin=6pt, labelsep=3pt]
\item We review three mitigation strategies for supplementing model-specific unknown knowledge, categorized by the extent of modification to the LLM's parameters. (\cf Figure~\ref{fig:mitigation_5.2}).
\setlength{\labelsep}{1pt}
\item[\idea] Future research could explore adaptive frameworks that integrate external retrieval with internal model updates for continuous knowledge improvement with minimal disruption.
\end{itemize}
\end{minipage}
} \\ [-2mm]
\subsection{Model-agnostic Unknown Knowledge}
In addressing model-agnostic unknown knowledge, two primary strategies, \textit{refusal} and \textit{asking clarification questions}, can be employed to ensure that LLMs respond appropriately.

\paragraph{Refusal}

Faced with queries involving model-agnostic unknown knowledge, LLMs are expected to refuse to answer for preventing misinformation. There are two primary methods for learning to refuse: \textit{prompt-based} and \textit{alignment-based} approaches.

Prompt-based approaches use designed prompts that help LLMs decide whether to refuse questions about unknown knowledge. The prompts are used to evaluate if a question involves unknown content to LLM \cite{DBLP:conf/emnlp/WenHW24, DBLP:conf/acl/AmayuelasWPCW24, agarwal2023nlpmodelsidentifydistinguish}, and to express the knowledge limitations \cite{DBLP:journals/corr/abs-2406-10881}. Also, LLMs can be prompted to justify their decision to decline a question \cite{Song2024MeasuringAE}.

Alignment-based approaches include supervised fine-tuning and reinforcement learning (RL) approaches. Supervised methods involve creating honesty alignment datasets, such as ``I don't know'' datasets, through instruct tuning to teach LLMs to admit uncertainty in responses \cite{Yang2023AlignmentFH, Cheng2024CanAA, Zhang2023RTuningIL, gao2024honestllm, zhu2025grait}. RL approaches generally constructs datasets that reflect user preferences, and use them to train LLMs through reward systems to discern when to refuse questions \cite{Cheng2024CanAA, Tomani2024UncertaintyBasedAI, Xu2024RejectionIR}.

\paragraph{Asking Clarification Questions}

When LLMs encounter questions involving model-agnostic unknown knowledge, asking clarification questions is an another common strategy. This method avoids direct uncertain responses and uses proactive dialogues to refine queries \cite{Deng2023ASO, Aliannejadi2021BuildingAE, Guo2021AbgCoQACA, Leippert2024ToCO, kim-etal-2024-aligning}. 
This is supported by specific prompt frameworks, with schemes encouraging LLMs to analyze questions deeply before responding \cite{Deng2023PromptingAE, Chen2024STYLEID}. Frameworks by \citet{kuhn2022clam} and \citet{Mu2023ClarifyGPTEL} enable LLMs to request clarifications selectively or identify unclear requirements, enhancing response accuracy. Latest methods like contrastive self-training and reward model learning help improve the quality of LLMs' questions in dialogues \cite{Chen2024LearningTC, Andukuri2024STaRGATETL}.

\vspace{2mm}
\noindent \colorbox{violet!8}{
\begin{minipage}{0.46\textwidth}
\ssmall
{\color{violet}{\textbf{Summary \& Ideas - Mitigation of Model-agnostic Unknown Knowledge}}}
\begin{itemize}[itemsep=-1pt, topsep=2pt, leftmargin=6pt, labelsep=3pt]
\item Refusal and asking clarification questions are two most widely-studied strategies for mitigating model-agnostic unknown knowledge.
\setlength{\labelsep}{1pt}
\item[\idea] Existing refusal strategies fail to differentiate between model-specific and model-agnostic unknown knowledge, leading to a degraded user experience when the query is, in fact, answerable.
\setlength{\labelsep}{1pt}
\item[\idea] There are certain issues about unintended side effects when inappropriately adopting these strategies, such as over-refusal and unnecessary cost.
\end{itemize}
\end{minipage}
} \\ [-2mm]

\section{Challenges and Prospects}\label{sec:challenge}

In this section, we discuss several significant challenges and emerging prospects along with the exploration of knowledge boundaries in LLMs.

\paragraph{Benchmark for Knowledge Boundary}
Various knowledge-based QA datasets are key benchmarks for assessing LLMs' knowledge boundaries, as summarized in Appendix \ref{app:dataset}. 
However, there are still critical areas lacking comprehensive benchmarks. 
Firstly, it lacks benchmarks for identifying the knowledge boundary of LLMs (\S \ref{sec:identification}). The benchmark construction should involve key aspects including multiple ground-truth answers, the influence of prompts, and reasoning complexity. Failing to answer a single question does not necessarily indicate whether the LLM can handle related knowledge \cite{DBLP:conf/acl/YinZR024}. 
Secondly, evaluating mitigation methods under different categories (\S \ref{sec:mitigation}) also requires corresponding benchmarks. 
A standardized benchmark is essential for enabling a thorough and fair comparison on the performance of various mitigation methods. 
Thus, our proposed taxonomy provides a systematic and valuable foundation to guide the development of these benchmarks.

\paragraph{Utilization of Knowledge Boundary}

Estimating and understanding LLMs' knowledge boundaries should not mark the end of the process. Instead, identifying these limitations can serve as a foundation for enhancing the model's performance in mitigating queries beyond their knowledge boundaries. For instance, the utilization of model uncertainty can reduce RAG costs and minimize the risk of introducing noise from external sources \cite{yao2024seakr}, or enhance the preference optimization by encouraging the LLM policy to differentiate reliable or unreliable feedback \cite{wang2024self}. 
Another instance is to enhance the robustness of LLMs against prompt sensitivity.
Some pioneer research study such issue regarding the order of demonstrations \cite{chen2025pearl, DBLP:conf/acl/LuBM0S22}. Further studies could investigate the role of outward knowledge boundary of LLMs in the overall prompt robustness of LLMs, enabling them to express more knowledge they already possess.

\paragraph{Understanding Knowledge Boundary through Knowledge Mechanisms}
Existing research on knowledge mechanisms, including memorization, comprehension, creation, and evolution, investigates how LLMs acquire, store, and utilize knowledge \cite{wang2024knowledge}. 
It is worth studying different phenomena of LLM knowledge boundaries under these mechanism views. For example, the outward knowledge boundary showcases how mechanisms like memorization and comprehension manifest in the explicit behaviors and outputs of the model, while the parametric boundary reflects a deeper, less visible level of how knowledge is embedded and structured through these mechanisms. The universal boundary can help measure the creative and evolutionary capabilities of LLMs.

\paragraph{Generalization of Knowledge Boundary}

While knowledge boundary studies are often conducted in specific domains, understanding the general knowledge boundary in LLMs is vital. 
The internal state probing approach has been validated with a certain generalization ability \cite{liu2024universal}, but it is still an open challenge whether trained probes can generalize well across domains as a general knowledge boundary detector, fostering refusal and input clarification in open domains. 
Further theoretical analysis and studies are needed to identify the existence and utility of general knowledge boundaries, which may be related to fundamental theories of LLM knowledge mechanism \cite{DBLP:conf/emnlp/WangYXQD00GJX0C24, DBLP:conf/icml/Allen-ZhuL24}.

\paragraph{Unintended Side Effects}
Although the mitigation strategies mentioned above aim to improve the performance of LLMs, they can also introduce a range of unintended side effects that may compromise the utility and effectiveness of the model. In the following, we detail several of these effects, highlighting the challenges and potential trade-offs.

\begin{itemize}[leftmargin=*]
    \item \textbf{Over-refusal} occurs when models excessively avoid responding, even to valid queries within their knowledge boundaries. Studies like \citet{Varshney2023TheAO} show that techniques like ``self-check'' can make LLMs overly cautious, reducing their utility. \citet{DBLP:journals/corr/abs-2410-06913} further explores this issue, identifying static and dynamic conflicts in training as key contributors.
    \item \textbf{Unnecessary Cost} arises when LLMs use strategies (\textit{e.g.}, clarifications, RAG, or self-correction) to manage queries beyond their knowledge boundaries. Although effective in avoiding undesired behaviors, these methods often consume additional time or effort, delaying responses. For instance, clarifications increase the round of interactions \cite{Chen2024STYLEID}, while RAG can introduce noise if LLMs already possess the necessary knowledge \cite{asai2024selfrag}. 
\end{itemize}

\paragraph{Knowledge Boundary in Long-Form Language Modeling}
Knowledge boundaries critically impact long-form factuality, defining how well LLMs generate coherent and accurate extended responses. Unlike short-form factuality, which depends on individual fact retrieval, long-form factuality is affected by cumulative knowledge gaps, where minor errors propagate over extended discourse. Existing research \citep{wei2024long, Min2023FActScoreFA, Huang2024FactAlignLF} has explored evaluation and mitigation strategies, providing a lens to examine how LLMs navigate and extend their knowledge boundaries, but the interaction between knowledge boundaries and factuality degradation remains an open research area.

\section{Conclusions}

This survey present a comprehensive overview of the knowledge boundary of LLMs, offering a formalized taxonomy and addressing key questions in the field. By exploring undesirable behaviors, identification techniques, and mitigation strategies, we emphasize the critical role of understanding and managing these boundaries to improve the reliability and utility of LLMs. Despite significant progress, challenges persist, including lack of comprehensive benchmarks,  potential uses of knowledge boundary, and the role of knowledge boundary under various mechanisms. We hope this survey inspires continued exploration and innovation toward more trustworthy and reliable LLMs.

\section*{Limitations}
We identify several limitations of our work. 

\paragraph{Formal Definition of Knowledge}
This survey does not give a formal definition of the knowledge $k$, which is a critical problem in the scope of NLP research on knowledge. 
In this survey, we define the abstracted concept of knowledge as $k$, which is represented by a set of textual expressions of input and output $Q_k$. 
This definition can facilitate practical NLP experiments and efficient validation. 
In fact, the formal definition of knowledge is still a debatable topic, calling for future exploration. 
For example, \citet{DBLP:conf/emnlp/FierroDSGS24} try to bridge the philosophical definition to the knowledge of LLMs, though significant disagreements persist among various philosophical schools of thought.

\paragraph{Various Forms of Textual Expressions regarding Different Knowledge Types}
Different types of knowledge may correspond to various forms of textual input-output $Q_k$, while we aim to provide a universal definition without the loss of generality. For instance, outputs for complex concepts may be open-ended and long-form, while simpler concepts might be expressed in a multiple-choice format. Some knowledge can be explicitly stated in the input, whereas others, such as commonsense knowledge, may need to be inferred from the input. Additionally, a single input may have multiple valid outputs. Some knowledge types, like mathematical knowledge, may inevitably involve multiple pieces of knowledge in a single input-output instance. As research progresses, a more nuanced definition for $Q_k$ may be necessary to accommodate different knowledge types effectively.

\paragraph{(Un)Known to Human or Models}
Besides, in our definition, LLMs operate within the universal knowledge boundary, typically limited to human-known knowledge. We generally believe that LLMs do not possess knowledge beyond this boundary. However, there may be outliers that LLMs have knowledge that is unknown for human, which is not clearly studied in existing research. \citet{DBLP:conf/emnlp/WangYXQD00GJX0C24} hypothesize that LLMs may create new knowledge, but its reliability remains uncertain. 
While such outputs could reflect meaningful discoveries, they may also stem from implicit correlations in training data. Since existing research has not systematically examined LLM-generated unknown knowledge, its nature and implications remain unclear.

\section*{Acknowledgment}
This research was supported by the Singapore Ministry of Education (MOE) Academic Research Fund (AcRF) Tier 1 grant (No. MSS24C012).

\bibliography{custom}

\appendix

\section{Overview}
\label{app:overview}

We begin by introducing the definition of knowledge boundary, outlining three types of knowledge boundaries and a four-type knowledge taxonomy. Following this, we describe the typical undesired behaviors that arise from knowledge limitations, emphasizing the importance of addressing such issues. These challenges highlight the critical need for methods that can detect when LLMs operate beyond their knowledge capabilities. To this end, we present three distinct identification techniques that help delineate where knowledge gaps exist. Once these gaps are identified, various mitigation strategies can be employed to address the issues caused by the knowledge boundaries. Finally, we explored several significant challenges and emerging prospects in understanding and managing knowledge boundaries in LLMs. Figure \ref{categorization_of_survey} illustrates a comprehensive framework for managing the knowledge boundaries of LLMs, focusing on three key components: Undesired Behaviors, Identification of Knowledge Boundaries, and Mitigation Strategies.

\tikzstyle{my-box}=[
    rectangle,
    draw=black,
    rounded corners,
    text opacity=1,
    minimum height=1.5em,
    minimum width=5em,
    inner sep=2pt,
    fill opacity=.5,
    text width=28em, 
    align=left, 
    execute at begin node=\strut 
]

\tikzstyle{behaviour_leaf}=[my-box, minimum height=1.5em,
    fill=cyan!20, text=black, align=left,font=\scriptsize,
    inner xsep=2pt,
    inner ysep=5pt,
]
\tikzstyle{identification_leaf}=[my-box, minimum height=1.5em,
    fill=lightgreen!20, text=black, align=left,font=\scriptsize,
    inner xsep=2pt,
    inner ysep=5pt,
]
\tikzstyle{mitigation_leaf}=[my-box, minimum height=1.5em,
    fill=lightpurple!20, text=black, align=left, font=\scriptsize,
    inner xsep=2pt,
    inner ysep=5pt,
]
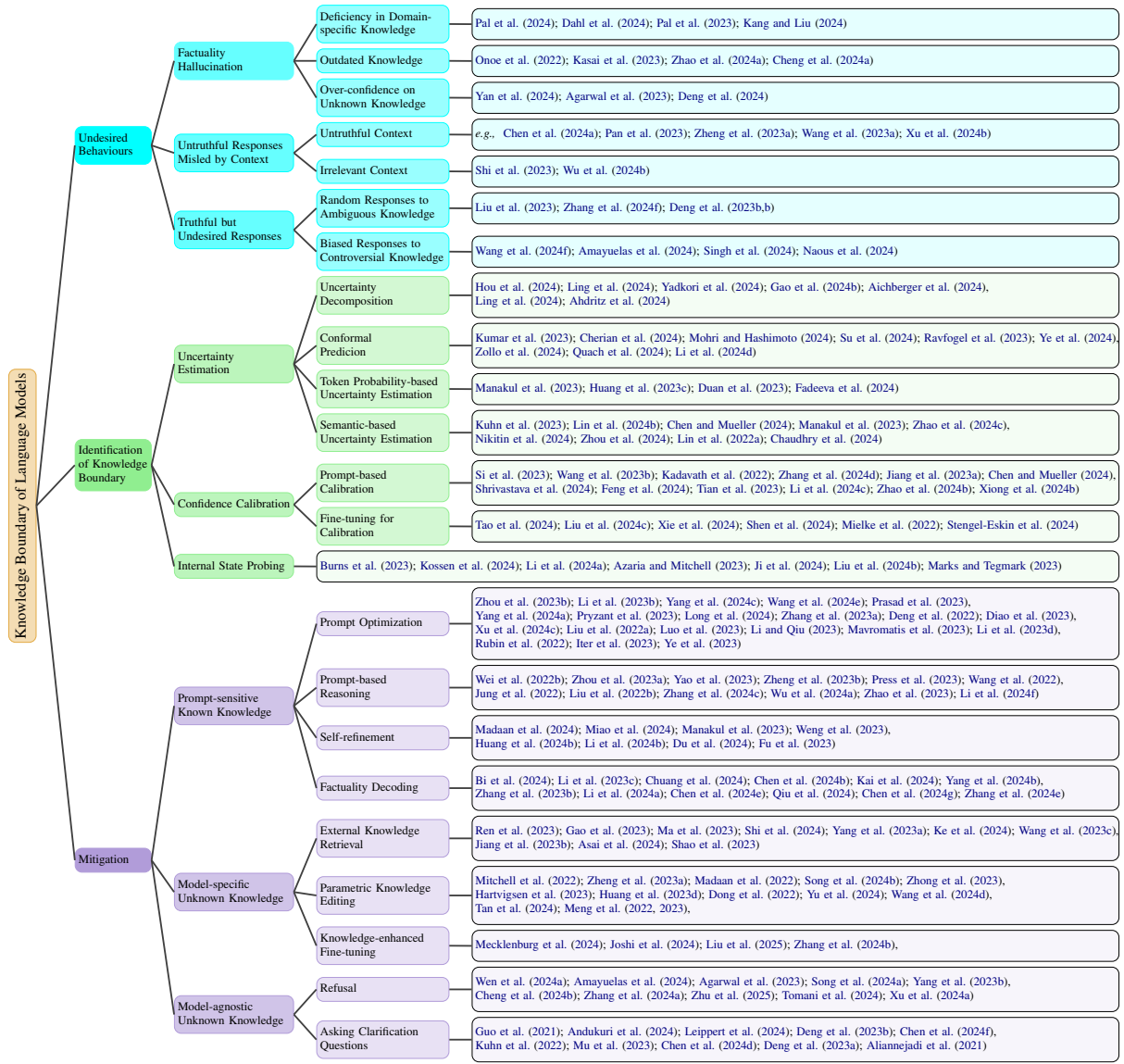
\begin{figure*}[tp]
    \centering
    \resizebox{\textwidth}{!}{
        \begin{forest}
            for tree={
                grow=east,
                reversed=true,
                anchor=base west,
                parent anchor=east,
                child anchor=west,
                base=left,
                font=\small,
                rectangle,
                draw=black,
                rounded corners,
                align=left,
                minimum width=4em,
                edge+={darkgray, line width=1pt},
                s sep=3pt,
                inner xsep=2pt,
                inner ysep=3pt,
                ver/.style={rotate=90, child anchor=north, parent anchor=south, anchor=center},
            },
            where level=1{text width=3.7em,font=\scriptsize,}{},
            where level=2{text width=6.0em,font=\scriptsize,}{},
            where level=3{text width=6.7em,font=\scriptsize,}{},
            [
                Knowledge Boundary of Language Models, ver, color=harvestgold!100, fill=harvestgold!40,
                text=black
                [
                    Undesired \\ Behaviours, color=cyan!100, fill=cyan!100, text=black
                    [
                        Factuality \\ Hallucination, color=cyan!60, fill=cyan!60, text=black
                        [
                            Deficiency in Domain-\\specific  Knowledge, color=cyan!100, fill=cyan!40, text=black
                            [
                                {\citet{pal2024domain, legal-hallucination, medical-hallucination, finance-hallucination}}
                                , behaviour_leaf, text width=34.2em
                            ]
                        ]
                        [
                            Outdated Knowledge, color=cyan!100, fill=cyan!40, text=black
                            [
                                {\citet{naacl22-outdated, kasai2024realtime, Zhao2024SetTC, Cheng2024DatedDT}}
                                , behaviour_leaf, text width=34.2em
                            ]
                        ]
                        [
                            Over-confidence on \\ Unknown Knowledge, color=cyan!100, fill=cyan!40, text=black
                            [
                                {\citet{yan2024reward, agarwal2023nlpmodelsidentifydistinguish, deng2024gotcha}}
                                , behaviour_leaf, text width=34.2em
                            ]
                        ]
                    ]
                    [
                        Untruthful Responses \\ Misled by Context, color=cyan!100, fill=cyan!60, text=black
                        [
                            Untruthful Context, color=cyan!100, fill=cyan!40, text=black
                            [
                                {\eg~\citet{chen2024editingllmsinjectharm, Pan2023OnTR, ike, wang2023can, Xu2023TheEI}}
                                , behaviour_leaf, text width=34.2em
                            ]
                        ]
                        [
                            Irrelevant Context, color=cyan!100, fill=cyan!40, text=black
                            [
                                {\citet{shi2023large, Wu2024HowED}}
                                , behaviour_leaf, text width=34.2em
                            ]
                        ]
                    ]
                    [
                        Truthful but \\ Undesired Responses, color=cyan!100, fill=cyan!60, text=black
                        [
                            Random Responses to\\ Ambiguous Knowledge, color=cyan!100, fill=cyan!40, text=black
                            [
                                {\citet{emnlp23-ambiguity,Zhang2024CLAMBERAB, Deng2023PromptingAE, Deng2023PromptingAE}}
                                , behaviour_leaf, text width=34.2em
                            ]
                        ]
                        [
                            Biased Responses to \\ Controversial  Knowledge , color=cyan!100, fill=cyan!40, text=black
                            [
                                {\citet{coling24-controversial,DBLP:conf/acl/AmayuelasWPCW24, Singh2024BornWA,acl24-culturalbias}}
                                , behaviour_leaf, text width=34.2em
                            ]
                        ]
                    ]
                ]
                [
                    Identification \\ of Knowledge \\ Boundary, color=lightgreen!100, fill=lightgreen!100, text=black
                    [
                        Uncertainty \\ Estimation, color=lightgreen!100, fill=lightgreen!60, text=black
                        [
                           Uncertainty \\ Decomposition, color=lightgreen!100, fill=lightgreen!40, text=black
                            [
                                {\citet{hou2023decomposing, ling2024uncertainty, yadkori2024believe, gao2024spuq, aichberger2024how}, \\ \citet{ling2024uncertainty, ahdritz2024distinguishing}}
                                , identification_leaf, text width=34.2em
                            ]
                        ]
                        [
                            Conformal \\ Predicion, color=lightgreen!100, fill=lightgreen!40, text=black
                            [
                                {\citet{DBLP:journals/corr/abs-2305-18404, DBLP:journals/corr/abs-2406-09714, DBLP:conf/icml/MohriH24, DBLP:conf/emnlp/SuLWC24, DBLP:conf/acl/RavfogelGG23, DBLP:conf/nips/0001YPWWY0T24}, \\ \citet{DBLP:conf/iclr/ZolloMDSPZ24, DBLP:conf/iclr/QuachFSYSJB24, DBLP:conf/naacl/Li00B24}}
                                , identification_leaf, text width=34.2em
                            ]
                        ]
                        [
                            Token Probability-based \\ Uncertainty Estimation, color=lightgreen!100, fill=lightgreen!40, text=black
                            [
                                {\citet{manakul2023selfcheckgpt, huang2023look, duan2024shifting,fadeeva2024fact}}
                                , identification_leaf, text width=34.2em
                            ]
                        ]
                        [
                            Semantic-based \\ Uncertainty Estimation, color=lightgreen!100, fill=lightgreen!40, text=black
                            [
                                {\citet{kuhn2023semantic, lin2024generating, chen2024quantifying, manakul2023selfcheckgpt, zhao2024knowing}, \\ \citet{nikitin2024kernel, zhou2024relying, lin2022teaching, chaudhry2024finetuning}}
                                , identification_leaf, text width=34.2em
                            ]
                        ]
                    ]
                    [
                       Confidence Calibration, color=lightgreen!100, fill=lightgreen!60, text=black
                        [
                            Prompt-based \\ Calibration, color=lightgreen!100, fill=lightgreen!40, text=black
                            [
                                {\citet{si2023prompting, wang2023selfconsistency, kadavath2022language, zhang2024calibrating, jiang2023calibrating, chen2024quantifying}, \\ \citet{shrivastava2024llamas, feng-etal-2024-dont, tian2023just, li2024think, zhao-etal-2024-fact, xiong2024can}}
                                , identification_leaf, text width=34.2em
                            ]
                        ]
                        [
                            Fine-tuning for \\ Calibration, color=lightgreen!100, fill=lightgreen!40, text=black
                            [
                                {\citet{tao-etal-2024-trust, liu2024litcab, xie-etal-2024-calibrating, shen2024thermometer, mielke2022reducing, stengel2024lacie}}
                                , identification_leaf, text width=34.2em
                            ]                        ]
                    ]
                    [
                        Internal State Probing, color=lightgreen!100, fill=lightgreen!60, text=black
                        [
                            {\citet{burns2022discovering, kossen2024semantic, li2024inference, azaria-mitchell-2023-internal, ji-etal-2024-llm, liu2024universal, marks2023geometry}
                            }
                            , identification_leaf, text width=42.5em
                        ]
                    ]
            ]
            [
                    Mitigation, color=lightpurple!100, fill=lightpurple!100, text=black
                    [
                        Prompt-sensitive \\ Known Knowledge, color=lightpurple!100, fill=lightpurple!60, text=black
                        [
                             Prompt Optimization, color=lightpurple!100, fill=lightpurple!30, text=black
                            [
                                {\citet{zhou2023large, li2023robust, yang-etal-2024-dual, wang2024promptagent, prasad2023grips}, \\  \citet{yang2024large, pryzant2023automatic, do2023prompt, zhang2023tempera, deng2022rlprompt, diao2023blackbox}, \\ \citet{xu2024context, liu2022makes, luo2023dricl, li2023finding, mavromatis2023examples, li2023unified}, \\ \citet{rubin2022learning, iter2023context, ye2023compositional}}
                                , mitigation_leaf, text width=34.2em
                            ]
                        ]
                        [
                            Prompt-based \\ Reasoning, color=lightpurple!100, fill=lightpurple!30, text=black
                            [
                                {\citet{wei2022chain, DBLP:conf/iclr/ZhouSHWS0SCBLC23, DBLP:conf/nips/YaoYZS00N23, DBLP:journals/corr/abs-2304-09797, press2023measuring, wang2022iteratively}, \\ \citet{ jung2022maieutic, liu2022generated, zhang2024tree, wu2024gendec, zhao2023verify, li2024chainofknowledge}
                                }
                                , mitigation_leaf, text width=34.2em
                            ]
                        ]
                        [
                            Self-refinement, color=lightpurple!100, fill=lightpurple!30, text=black
                            [
                                {\citet{madaan2024self, miao2024selfcheck, manakul2023selfcheckgpt, weng2023large}, \\ \citet{ huang2024large, li2024confidence, du2024improving, fu2023improving}}
                                , mitigation_leaf, text width=34.2em
                            ]
                        ]
                        [
                            Factuality Decoding, color=lightpurple!100, fill=lightpurple!30, text=black
                            [
                                {\citet{bi2024factuality, li2023contrastive, chuang2024dola, chen2024lower, kai2024sh2, yang2024improving}, \\ \citet{ zhang2023alleviating, li2024inference, chen2024context, qiu2024spectral, chen2024truth, zhang2024truthx}}
                                , mitigation_leaf, text width=34.2em
                            ]
                        ]
                    ]
                    [
                        Model-specific \\ Unknown Knowledge, color=lightpurple!100, fill=lightpurple!60, text=black
                        [
                            External Knowledge \\ Retrieval, color=lightpurple!100, fill=lightpurple!30, text=black
                            [
                                {\citet{ren2023investigating, gao-etal-2023-precise, ma-etal-2023-query, shi-etal-2024-replug, yang-etal-2023-prca, ke-etal-2024-bridging, wang2023learning}, \\ \citet{jiang-etal-2023-active, asai2024selfrag, shao-etal-2023-enhancing}}
                                , mitigation_leaf, text width=34.2em
                            ]
                        ]
                        [
                            Parametric Knowledge \\ Editing, color=lightpurple!100, fill=lightpurple!30, text=black
                            [
                                {\citet{serac, ike, memprompt, bbox-edit, mello}, \\ \citet{ grace, t-patcher, calinet, melo, wise}, \\ \citet{malmen, rome, memit},}
                                , mitigation_leaf, text width=34.2em
                            ]
                        ]
                        [
                            Knowledge-enhanced \\ Fine-tuning, color=lightpurple!100, fill=lightpurple!30, text=black
                            [
                                {\citet{Mecklenburg2024InjectingNK, Joshi2024AdaptingML,Liu2024StructureawareDK, Zhang2024SyntheticKI},}
                                , mitigation_leaf, text width=34.2em
                            ]
                        ]
                    ]
                    [
                        Model-agnostic \\ Unknown Knowledge, color=lightpurple!100, fill=lightpurple!60, text=black
                        [
                            Refusal, color=lightpurple!100, fill=lightpurple!30, text=black
                            [
                                {\citet{DBLP:conf/emnlp/WenHW24, DBLP:conf/acl/AmayuelasWPCW24, agarwal2023nlpmodelsidentifydistinguish, Song2024MeasuringAE, Yang2023AlignmentFH}, \\ \citet{Cheng2024CanAA, Zhang2023RTuningIL, zhu2025grait, Tomani2024UncertaintyBasedAI, Xu2024RejectionIR}}
                                , mitigation_leaf, text width=34.2em
                            ]
                        ]
                        [
                            Asking Clarification \\ Questions, color=lightpurple!100, fill=lightpurple!30, text=black
                            [
                                {\citet{ Guo2021AbgCoQACA,
                                Andukuri2024STaRGATETL, Leippert2024ToCO,Deng2023PromptingAE,Chen2024STYLEID}, \\ \citet{kuhn2022clam,Mu2023ClarifyGPTEL,Chen2024LearningTC,Deng2023ASO,Aliannejadi2021BuildingAE}}
                                , mitigation_leaf, text width=34.2em
                            ]
                        ]
                    ]
                ]
        ]
        ]
        \end{forest}
    }
    \caption{The main content flow and categorization of this survey.}
    \label{categorization_of_survey}
\end{figure*}

\paragraph{Summary of Contribution}
As a survey paper, our primary goal is to synthesize and analyze existing research while providing new insights and frameworks for understanding the knowledge boundaries of LLMs. We believe our work offers significant novelty in the following aspects: 
\paragraph{1) Scope and Coverage}
\begin{itemize}[leftmargin=*]
    \item \textbf{Novelty in Scope}: This survey covers a topic or area that has not been thoroughly reviewed before. We address an emerging field and underexplored topics.
    \item \textbf{Comprehensiveness}: This survey provides a comprehensive or up-to-date overview. Novelty lies in including recent advancements, overlooked studies, or a broader range of perspectives.
\end{itemize}
\paragraph{2) Organization and Structure}
\begin{itemize}[leftmargin=*]
    \item \textbf{Unique Frameworks or Taxonomies}: This survey introduces a novel taxonomy, as a new way of categorizing, organizing, or analyzing the literature of LLM knowledge boundary studies. This new taxonomy provides fresh insights into the field.
\end{itemize}
\paragraph{3) Insights and Critical Analysis}
\begin{itemize}[leftmargin=*]
    \item \textbf{Original Insights}: This survey provides original interpretations and thought-provoking perspectives on the existing literature. For instance, we provide a unique categorization of mitigation strategies for prompt-sensitive known knowledge according to the process on the LLM input and output sides, and also for model-specific unknown knowledge based on the extent of modification to LLM's parameters.
    \item \textbf{Identification of Gaps}: This survey identifies several underexplored areas or open problems in the field.
\end{itemize}
\paragraph{4) Timeliness and Relevance}
\begin{itemize}[leftmargin=*]
    \item As  \textbf{the very first survey paper} on the topic of LLM knowledge boundaries, our work serves as a foundational resource for researchers and practitioners. By consolidating and organizing the existing literature, we provide a starting point for further exploration and innovation in this critical area.
\end{itemize}
In summary, our work contributes novelty through comprehensive coverage, innovative taxonomy, and original insights. We believe these contributions significantly advance the understanding of LLM knowledge boundaries and provide a valuable resource for the research community.

\section{Dataset}
\label{app:dataset}
\begin{table*}[ht!]
\tiny
\centering
\begin{tabular}{
  p{2cm}  
  p{1.8cm}
  p{2.3cm} 
  p{0.4cm} 
  p{6cm}   
}
\toprule
 \textbf{Knowledge Category} & 
 \textbf{Dataset} & 
 \textbf{Reference} & 
 \textbf{Size} &
 \textbf{Description} \\ 
 \midrule
 \multirow{30}{*}{\parbox{3.3cm}{\textbf{Prompt-Sensitive \\ Known Knowledge}}}
    & ProntoQA & \citet{PrOntoQA}   & 9.7k & A question-answering dataset which generates examples with chains-of-thought that describe the reasoning required to answer the questions correctly. \\ 
    \cmidrule{2-5} 
    & 2WikiMultiHopQA & \citet{ho2020constructing}  & 192,606 & A multi-hop QA benchmark combining structured and unstructured data.\\ 
    \cmidrule{2-5} 
    & MuSiQue & \citet{trivedi2022musique}   & 25k & A multi-hop QA benchmark with 2-4 hop questions.  \\ 
    \cmidrule{2-5} 
    & HotpotQA & \citet{yang2018hotpotqa}   & 113k & A multi-hop QA dataset requiring reasoning over two Wikipedia paragraphs, with supporting facts provided for explainability and evaluation. \\
    \cmidrule{2-5} 
    & TruthfulQA & \citet{lin2022truthfulqa}   & 817 & A benchmark across 38 categories, designed to evaluate whether language models generate truthful answers, particularly in cases prone to false beliefs. \\
    \cmidrule{2-5} 
    & PARAREL & \citet{elazar2021measuring}   & 328 & A dataset of English cloze-style query paraphrases for 38 relations, designed to evaluate the consistency of PLMs in handling factual knowledge across meaning-preserving input variations. \\
    \cmidrule{2-5} 
    & KAssess & \citet{dong2024statistical}   & 139k    & A comprehensive assessment suite with 994,123 entities and 600 relations, designed to evaluate the factual knowledge of LLMs by estimating their ability to generate correct answers across diverse prompts compared to random chance. \\
    \cmidrule{2-5} 
    & FARM & \citet{Xu2023TheEI}   & 1,952 & A dataset of factual questions paired with systematically generated persuasive misinformation, designed to evaluate the susceptibility of LLMs to belief manipulation through multi-turn persuasive conversations. \\
    \cmidrule{2-5} 
    & Misinfo-QA & \citet{Pan2023OnTR}   &  3,034 & A dataset designed to study the impact of misinformation on open-domain question answering (ODQA) systems by injecting synthetic misinformation passages to evaluate how QA models respond under such conditions. \\
    \cmidrule{2-5} 
\cmidrule{1-5}
        
\multirow{40}{*}{\parbox{3.3cm}{\textbf{Model-Specific \\ Unknown Knowledge}}}   
    & Natural Questions & \citet{kwiatkowski2019natural}   & 7,842 & A large-scale dataset of real anonymized Google queries, annotated with long and short answers from Wikipedia or marked null if no answer is present. \\
    \cmidrule{2-5} 
    & TopiOCQA & \citet{adlakha-etal-2022-topiocqa}   & 3,920 & An open-domain conversational dataset with information-seeking conversations featuring topic switches. \\ 
    \cmidrule{2-5}   
    & PopQA & \citet{mallen2022not}  & 14k & Long-tail relation triples from WikiData are converted into QA pairs; no explicit unanswerable questions but questions are about long-tail entities. \\ 
    \cmidrule{2-5}   
    & TriviaQA & \citet{joshi2017triviaqa}   &  950k & A realistic text-based question answering dataset which includes question-answer pairs from documents collected from Wikipedia and the web. \\ 
    \cmidrule{2-5}        
    & RealtimeQA & \citet{kasai2024realtime}   & 4,356 & A dynamic open-domain question-answering dataset that evaluates models based on real-time, time-sensitive questions sourced weekly from news articles. \\ 
    \cmidrule{2-5}   
    & FreshQA & \citet{vu2023freshllmsrl}  & 600 & A dynamic QA benchmark designed to evaluate LLMs on fast-changing world knowledge and debunking false premises. \\ 
    \cmidrule{2-5} 
    & PubMedQA & \citet{Jin2019PubMedQAAD}  & 273.5k & A biomedical research question-answering dataset, which features questions derived from research article titles in PubMed, requiring complex reasoning and interpretation of quantitative biomedical content. \\
    \cmidrule{2-5} 
    & MIRAGE & \citet{Xiong2024BenchmarkingRG} & 7,663 & A benchmark dataset for medical question answering, focusing on retrieving information from medical literature to answer multiple-choice medical questions, with an emphasis on zero-shot reasoning and systematic evaluation of retrieval performance. \\
    \cmidrule{2-5} 
    & TAT-QA & \citet{Zhu2021TATQAAQ}  & 16,552 & A question-answering dataset for the financial domain, combining tabular and textual content from real financial reports. \\
    \cmidrule{2-5} 
    & FinQA & \citet{Chen2021FinQAAD}  & 8,281 & A question-answering dataset for the financial domain, with questions and answers crafted by financial experts, involving complex numerical reasoning over tables and text from financial reports. \\
    \cmidrule{2-5} 
    & JEC-QA & \citet{Zhong2019JECQAAL} & 26,365 & A legal-domain question-answering dataset with questions sourced from the National Judicial Examination of China, covering legal concept understanding and case analysis. \\
    \cmidrule{2-5} 
    & LawBench & \citet{Fei2024LawBenchBL} & 20,000 & A legal reasoning evaluation benchmark designed for the Chinese legal environment, covering tasks such as legal knowledge memorization, document proofreading, case analysis, charge prediction, and legal consultation. \\
\cmidrule{1-5}
 
\multirow{15}{*}{\parbox{3.3cm}{\textbf{Model-Agnostic \\ Unknown Knowledge}}} 
    & KUQ & \citet{DBLP:conf/acl/AmayuelasWPCW24}   & 6,884 & A dataset designed to explore uncertainty in question-answering by focusing on questions without definitive answers. \\ 
        \cmidrule{2-5}   
    & UnknownBench & \citet{liu2024examining}  & 13,319 & A benchmark consisting of answerable and unanswerable questions, designed to evaluate LLMs' ability to express uncertainty and handle knowledge gaps while maintaining honesty and helpfulness. \\ 
        \cmidrule{2-5}  
    & SelfAware & \citet{DBLP:conf/acl/YinSGWQH23}   & 2,337 & A dataset containing unanswerable questions across five categories, designed to evaluate LLMs' self-knowledge by detecting uncertainty and their ability to identify limitations in their knowledge. \\ 
        \cmidrule{2-5}  
    & QnotA & \citet{agarwal2023nlpmodelsidentifydistinguish}  & 400  &  A dataset featuring questions without definitive answers across five categories, paired with corresponding answerable alternatives. \\ 
        \cmidrule{2-5}  
    & KUQP & \citet{deng2024gotcha}   & 320 & A dataset of known and unknown question pairs, designed to evaluate language models' ability to handle unanswerable, ambiguous, or incorrect queries. \\ 
        
      \midrule
\bottomrule
\end{tabular}
\caption{Representative datasets for studying the knowledge boundary of language models.} 
\label{tab:all_benchmarks}
\end{table*}

In the pursuit of advancing LLM capabilities and understanding their boundaries in knowledge processing, various datasets have been meticulously designed and utilized. 
The following sections categorize these datasets into three distinct groups based on the type of knowledge they aim to verify: Prompt-Sensitive Known Knowledge, Model-Specific Unknown Knowledge, and Model-Agnostic Unknown Knowledge. A summary of these datasets can be viewed in Table~\ref{tab:all_benchmarks}.

\paragraph{Datasets for Prompt-Sensitive Known Knowledge}

This type of datasets mainly aim to assess the prompt-sensitive known knowledge of LLMs, requiring specific prompting strategies and decoding strategies for the LLM to fully recall and utilize such knowledge.

The first type of datasets focuses on \textbf{\emph{multi-step reasoning}}, such as multi-step knowledge-based question answering datasets (\textit{e.g.}, 2WikiMultiHopQA \cite{ho2020constructing}, MuSiQue \cite{trivedi2022musique}, and HotpotQA \cite{yang2018hotpotqa}) and logical reasoning datasets like ProntoQA \cite{PrOntoQA}. These tasks require the LLM to achieve a step-by-step reasoning process or benefit from prompting strategies that focus on question decomposition and explicit knowledge recall.

The second type is \textbf{\emph{fact-based question answering}} datasets that evaluate the LLM's factuality, \eg TruthfulQA \cite{lin2022truthfulqa}. In these datasets, the decoding strategy can influence how accurately knowledge is expressed \cite{li2024inference}. 

The third type of datasets explicitly study the influence of \textbf{\emph{varied prompt phrasing}} in LLM knowledge, including PARAREL \cite{elazar2021measuring} and KAssess \cite{dong2024statistical}. 

The fourth type involves datasets with \textbf{\emph{misleading contexts}}. 
\citet{wang2023can} curate queries with misleading user opinion to test LLM's ability to defend its response. 
FARM \cite{Xu2023TheEI} contains persuasive misinformation in the dialog context to evaluate LLM's belief change. 
Misinfo-QA \cite{Pan2023OnTR} includes model-generated misinformation to perturb open-domain QA.

\paragraph{Dataset for Model-Specific Unknown Knowledge}

This type of datasets can be used for assessing the model-specific unknown knowledge of LLMs, which challenges LLMs by probing their ability to handle highly specialized and temporally-sensitive information, testing their adaptive knowledge boundaries. These datasets are specifically designed to evaluate knowledge that lies outside the parametric scope of LLMs, requiring external knowledge retrieval or new knowledge injection to generate accurate responses.

Open-domain question answering datasets form an important category. These datasets evaluate the ability of language models to answer questions across a broad range of domains, leveraging both retrieval and parametric knowledge. Representative examples include Natural Questions \cite{kwiatkowski2019natural}, TopiOCQA (\citealt{adlakha-etal-2022-topiocqa}), PopQA (\citealt{mallen2022not}), and TriviaQA-unfiltered (\citealt{joshi2017triviaqa}). These datasets often focus on queries that require world knowledge or niche details, testing the model’s capacity to combine retrieval and internalized knowledge effectively. 
Meanwhile, various domain-specific QA datasets can be adopted to evaluate the model-specific unknown knowledge for each specialized applications, such as medical domain (\textit{e.g.}, PubMedQA \cite{Jin2019PubMedQAAD} and MIRAGE \cite{Xiong2024BenchmarkingRG}), finance domain (\textit{e.g.}, TAT-QA \cite{Zhu2021TATQAAQ} and FinQA \cite{Chen2021FinQAAD}) , and legal domain (\textit{e.g.}, JEC-QA \cite{Zhong2019JECQAAL} and LawBench \cite{Fei2024LawBenchBL}).

Another crucial subdomain focuses on time-sensitive datasets that test a model’s ability to generalize to out-of-distribution data. Datasets such as RealtimeQA (\citealt{kasai2024realtime}) and FreshQA(\citealt{vu2023freshllmsrl}) require language models to stay current with global events and provide accurate, up-to-date responses. These datasets evaluate the model’s capacity to adapt to evolving information and address queries that rely on recent developments.

This diverse set of datasets for studying model-sensitive unknown knowledge systematically evaluates the gaps in parametric knowledge of language models, testing their ability to retrieve, adapt, and reason with external information under various constraints.

\paragraph{Dataset for Model-Agnostic Unknown Knowledge}

As for the model-agnostic unknown knowledge, datasets such as Known-Unknown Questions (KUQ) \cite{DBLP:conf/acl/AmayuelasWPCW24} and UnknownBench \cite{liu2024examining} are specifically crafted to probe questions that remain unresolved or are based on uncertain future developments and incorrect assumptions. These datasets encapsulate complex scenarios including counterfactuals and ambiguities, which emphasize the current boundaries of our knowledge and the unpredictable nature of future inquiries.

Further pushing these boundaries, the SelfAware dataset \cite{DBLP:conf/acl/YinSGWQH23} explores questions that defy scientific consensus, are subjective, or philosophical, often requiring responses that extend beyond factual representation and into personal belief or theoretical speculation. Similarly, resources like QnotA \cite{agarwal2023nlpmodelsidentifydistinguish} and Known-Unknown Question Pairs (KUQP) \cite{deng2024gotcha} challenge models with incomplete or erroneous information and speculative predictions about the future. These datasets collectively serve to test LLM's capability in navigating the complexities of human inquiry where the answers are unknown.

\section{Details in Mitigation Approaches}

\subsection{Prompt-Sensitive Known Knowledge}

\paragraph{Prompt Optimization.}
For instruction optimization, 
APE \cite{zhou2023large} leverages LLMs to automatically forward-generate and perform Monte Carlo search on the prompts, and evaluate the performance of the candidate prompts via reverse generation, which consists of $n$ rounds. 
For demonstration optimization, 
KATE \cite{liu2022makes} retrieve the K nearest in-context examples by the semantic similarity to the test example, measured by the embedding from an encoder model. 

\paragraph{Prompt-based Reasoning.}
Chain-of-thoughts \cite{wei2022chain} generates the step-by-step rationales followed by the answer. 
Tree-of-thoughts \cite{DBLP:conf/nips/YaoYZS00N23} improves the linear chain-of-thoughts reasoning into tree structure, each node representing a piece of thoughts, and branches represents alternative thoughts. It allows LLMs to perform various forms of reasoning steps. 
Progressive-hint-prompting \cite{DBLP:journals/corr/abs-2304-09797} appends the LLM-generated answers to the prompt as hints to iteratively arrive at the correct answers. 

\paragraph{Self-refinement.}
Self-refine \cite{madaan2024self} prompts LLMs to generate feedback on its previous answer for iterative answer refinement. 
Self-verification \cite{weng2023large} transforms the generated answer into abductive reasoning questions to examine the consistency with the given context. 
Self-correction \cite{huang2023large} employs an interative initial CoT, review, and answer improvement process. 
MAD \cite{du2024improving} utilize multiple LLM agents to evaluate other LLMs' answers and update their own answers until they reach a consensus. 

\paragraph{Factuality Decoding.}
DoLA \cite{chuang2024dola} contrasts the logits obtained from the later layers with that obtained from the earlier layers to reduce generating factual errors. 
ITI \cite{li2024inference} changes the direction of the activations towards a factuality-improving direction obtained via probing to enhance factuality during inference. 

\subsection{Model-specific Unknown Knowledge}

\paragraph{External Knowledge Retrieval} For pre-generation methods, HyDE \cite{gao-etal-2023-precise} enhance retrieval by rewriting or expanding the user's input to obtain more comprehensive and accurate relevant information required by the model. This approach focuses on adapting the query to improve retrieval performance. For on-demand methods, FLARE \cite{jiang-etal-2023-active} evaluates the confidence levels in the model's generated content and actively retrieves pertinent documents to regenerate low-confidence segments, enhancing factual accuracy.

\paragraph{Parametric Knowledge Editing} 
PostEdit \cite{bbox-edit} edits the outputs of black-box LLMs while preserving data privacy and maintaining the original text style through fine-grained modifications. MELO \cite{melo} dynamically activates LoRA blocks using a neuron-indexed vector database, enabling efficient and precise updates to LLMs with minimal computational cost.

\paragraph{Knowledge-enhanced Fine-tuning} 
Fact-based SFT \cite{Mecklenburg2024InjectingNK} constructs a systematically covered fact-level question-answer dataset by extracting key facts from documents and generating diverse training examples, then enhances LLMs through SFT to improve their accuracy and adaptability to out-of-domain knowledge.
StructTuning \cite{Liu2024StructureawareDK} constructs structured domain knowledge by automatically extracting knowledge taxonomies from corpora, linking text segments to specific knowledge points for efficient model fine-tuning.
Factuality alignment methods \cite{DBLP:conf/nips/LinGOXLY024, DBLP:conf/emnlp/HuangC24a, DBLP:conf/iclr/TianMYMF24} is also a category of approach under this type, enhancing LLM knowledge via alignment approaches such as DPO. 

\subsection{Model-agnostic Unknown Knowledge} 

\paragraph{Refusal} \citet{DBLP:conf/acl/AmayuelasWPCW24} guides LLMs to recognize ``known-unknown'' questions and express uncertainty in high-uncertainty scenarios, enabling them to refrain from answering questions lacking definitive answers.
R-tuning \cite{Zhang2023RTuningIL} identifies the gap between the knowledge contained in the dataset and the knowledge encapsulated in the pre-trained parameters, thereby constructing a refusal-aware dataset and training the model based on it.

\begin{table*}[!t]
\small
\resizebox{\textwidth}{!}{
\begin{tabular}{l|p{4cm}|p{4cm}|p{4cm}}
\toprule
Type                                 & Method          & Training Cost      & Inference Cost                                                                     \\
\midrule
Prompt Optimization                  & APE \cite{zhou2023large}  &N/A           & $n$ round (prompt forward generation/monte carlo search + prompt reverse generation) \\ \midrule
\multirow{2}{*}{Prompt-based Reasoning}               & CoT \cite{NEURIPS2022_9d560961}   &N/A          & step-by-step reasoning + answer                                                           \\
                                     & PHP \cite{DBLP:journals/corr/abs-2304-09797}       &N/A      & $n$ round * (step-by-step reasoning + answer)                                              \\ \midrule
\multirow{2}{*}{Self-refinement}                      & Self-correction \cite{huang2024large}  & N/A & initial generation + review + revise                                         \\
                                     & MAD  \cite{du2024improving}  &N/A          & $n$ round * $m$ agent                                                          \\ \midrule
\multirow{2}{*}{Factuality Decoding} & DoLA \cite{chuang2024dola}  &N/A          & initial decoding + contrastive decoding step                                 \\
                                     & ITI \cite{li2023inferencetime}             & probing the truthful direction & perturbed attention decoding step          \\ \midrule
\multirow{2}{*}{External Knowledge Retrieval}   & HyDE \cite{gao-etal-2023-precise}        & N/A & hypothetical document generation + retrieval + generation                                     \\ 
                                                & FLARE \cite{jiang-etal-2023-active}   & N/A & $n$ * (retrieval + generation)                                  \\ \midrule
\multirow{2}{*}{Parametric Knowledge Editing}   & postEdit \cite{bbox-edit}     & retrieval + SFT  & generation                                        \\ 
                                                & MELO \cite{melo}          & retrieval + SFT & generation                               \\ \midrule
\multirow{2}{*}{Knowledge-enhanced Fine-tuning}                  &  Fact-based SFT \cite{Mecklenburg2024InjectingNK}     & fact extraction +  SFT & generation                  \\ & StructTuning \cite{Liu2024StructureawareDK}           &  structure-aware continual pre-training +  SFT & generation                                      \\ \midrule
\multirow{2}{*}{Refusal}                        & KUQ \cite{DBLP:conf/acl/AmayuelasWPCW24}   & SFT & generation                    \\ 
                                                & R-tuning \cite{Zhang2023RTuningIL}          & SFT & generation                                  \\ \midrule
\multirow{2}{*}{Asking Clarification Questions} & ProCoT \cite{Deng2023PromptingAE} & N/A & step-by-step reasoning + generation                                            \\ 
                                                & ACT \cite{Chen2024LearningTC}    & preference data construction + direct preference optimization & $n$ * generation \\
\bottomrule
\end{tabular}
}
\caption{Cost-effective comparison of representative mitigation techniques. }
\label{tab:cost_efective}
\end{table*}

\paragraph{Asking Clarification Questions}

\citet{Deng2023PromptingAE} constructed a proactive prompting scheme for dialogue between users and LLMs, requiring LLMs to carefully analyze and think through the question before posing clarification questions. ACT \cite{Chen2024LearningTC} guides the model to optimize dialogue strategies through contrastive learning in multi-turn conversations, especially when facing ambiguous user requests, enabling it to automatically recognize and ask clarification questions instead of guessing user intent or providing incorrect answers.

\section{Cost-effective Summarization of Representative Mitigation Techniques}
We present a cost-effective comparison of representative mitigation techniques in Section~\ref{sec:mitigation}, aiming to compare their usefulness and provide recommendations, as summarized in Table~\ref{tab:cost_efective}. This table offers a clearer  and more structured comparison of these methods, helping readers better understand their relative strengths and limitations. However, directly and fairly comparing the exact performance of these methods remains challenging due to the current lack of a general and comprehensive benchmark for evaluating different mitigation approaches. Further discussions on this challenge can be found in Section~\ref{sec:challenge}. 

From the table, we can make the following observations:
(1) Prompt optimization, prompt-based reasoning, and self-refinement typically follow two main patterns: step-by-step reasoning and multi-round refinement. These fundamental approaches enhance performance, though their specific design and cost vary depending on the method used.
(2) In factuality decoding, DoLA operates purely as a decoding method, whereas ITI includes a probing stage with parameter updates. This distinction can guide the choice between the two methods.
(3) The main frameworks of external knowledge retrieval and parametric knowledge editing focus on integrating retrieval and inference while minimizing the cost of both components.
(4) The cost of refusal and asking clarification questions methods mainly depends on whether fine-tuning on a constructed dataset is required.

\end{document}